\documentclass{article}
\usepackage[preprint]{neurips_2024}
\usepackage[utf8]{inputenc}
\usepackage[T1]{fontenc}
\usepackage{hyperref}
\usepackage{url}
\usepackage{amsfonts}
\usepackage{nicefrac}
\usepackage{tcolorbox}
\usepackage{tabularx}
\usepackage{booktabs}
\usepackage{microtype}
\usepackage{amsmath}
\usepackage{graphicx}
\usepackage{enumitem}
\usepackage{natbib}
\usepackage{algorithm}
\usepackage{algpseudocode}
\usepackage{float} 
\usepackage{caption}
\usepackage{times} 
\usepackage{adjustbox}

\title{SWE-Bench-CL: \\ Continual Learning for Coding Agents}

\author{
Thomas Joshi \\ Columbia University \\ \texttt{ttj2108@columbia.edu} \\
\And
Shayan Chowdhury \\ Columbia University\\ \texttt{sc4040@columbia.edu}
\And
Fatih Uysal \\ Columbia University\\ \texttt{fu2137@columbia.edu} \\
}

\begin{document}
\maketitle

\begin{abstract}
Large Language Models (LLMs) have achieved impressive results on static code‐generation benchmarks, but real‐world software development unfolds as a continuous stream of evolving issues, fixes, and feature requests. We introduce \textbf{SWE‑Bench‑CL}, a novel continual learning benchmark built on the human‑verified SWE‑Bench Verified dataset \cite{SWEBenchVerified}. By organizing GitHub issues into chronologically ordered sequences that reflect natural repository evolution, SWE‑Bench‑CL enables direct evaluation of an agent’s ability to accumulate experience, transfer knowledge across tasks, and resist catastrophic forgetting. We complement the dataset with (i) a preliminary analysis of inter‑task structural similarity and contextual sensitivity, (ii) an interactive, LangGraph‑based evaluation framework augmented with a FAISS‑backed semantic memory module, and (iii) a suite of specialized continual learning metrics—including average accuracy, forgetting, forward/backward transfer, tool‑use efficiency, and a generalized novel Composite Continual-Learning-Score and CL‑F$_\beta$ score—to capture the stability‐plasticity trade‑off. We outline a rigorous experimental protocol comparing memory‑enabled and memory‑disabled agents across diverse Python repositories. All code and data are publicly available at \url{https://github.com/thomasjoshi/agents-never-forget/}, providing the community with a reproducible platform for developing more adaptive and robust AI agents in software engineering.
\end{abstract}

\section{Introduction \& Motivation}
\label{sec:introduction}

Large Language Models (LLMs) have achieved remarkable success in a variety of code-related tasks, from autocompletion to generating entire code snippets from natural language descriptions \citep{chen2021evaluatinglargelanguagemodels,nijkamp2023codegenopenlargelanguage}. However, the lifecycle of real-world software projects is inherently dynamic and continuous. Repositories evolve daily: APIs are deprecated, libraries are upgraded, new bugs are discovered and fixed, and novel features are constantly requested. An adept software engineering agent must therefore not only generate correct code for an immediate request but also learn from its experiences, adapt to changes in the codebase, and, crucially, retain knowledge of how to handle past issues as the project grows and shifts. A human software engineer who has resolved 100 bugs in a complex codebase will be more adept at the 101st bug than an engineer new to it. This ability to accumulate experience is critical for \textbf{agents that continuously learn}.

However, current leading benchmarks for code intelligence—such as CodeSearchNet \citep{husain2020codesearchnetchallengeevaluatingstate}, CodeXGLUE \citep{lu2021codexgluemachinelearningbenchmark}, and even the robust SWE-Bench dataset \citep{jimenez2024swebenchlanguagemodelsresolve}---primarily evaluate models on isolated, static tasks. They typically present data as an unordered collection, lacking the temporal or sequential structure necessary to measure critical continual learning (CL) properties like adaptation to evolving contexts, knowledge retention over time, or the mitigation of catastrophic forgetting. These benchmarks often require only one-step retrieval or generation and employ evaluation metrics (e.g., BLEU, Exact Match, pass@k) that do not quantify an agent's ability to learn continuously or transfer knowledge effectively across related tasks. Furthermore, as we will demonstrate (Section~\ref{sec:harness_challenges}), evaluating sequentially structured, derived benchmarks like SWE-Bench-CL with harnesses designed for their static predecessors presents significant alignment challenges, further motivating the need for evaluation frameworks specifically designed for continual learning agents.

To bridge this significant gap, we introduce \textbf{SWE-Bench-CL}, a continual learning reformulation of the human-verified SWE-Bench Verified dataset \cite{SWEBenchVerified}—which itself is a refinement of the original SWE-Bench dataset \cite{jimenez2024swebenchlanguagemodelsresolve}. SWE-Bench-CL structures software engineering tasks (GitHub issues) from various repositories into chronologically ordered sequences, each designed to simulate a developer's ongoing engagement with a project. This temporal structuring allows for the direct assessment of an agent's ability to learn from a stream of tasks, adapt to new problems, and remember past solutions.

This paper makes the following primary contributions:
\begin{enumerate}[nosep,leftmargin=*]
    \item \textbf{A Novel Benchmark Dataset (SWE-Bench-CL):} In Section~\ref{sec:swe_bench_cl_benchmark}, we detail the construction and structure of SWE-Bench-CL, a reproducible, temporally organized benchmark designed to measure adaptation and memory retention in coding agents.
    \item \textbf{Preliminary Dataset Analysis:} In Section~\ref{sec:preliminary_dataset_analysis}, we present an analysis of SWE-Bench-CL's structural characteristics, including inter-task similarity and contextual sensitivity. These findings highlight the unique challenges the benchmark poses for continual learning and inform the design of effective evaluation strategies and agent architectures.
    \item \textbf{A Proposed Agentic Evaluation Framework:} In Section~\ref{sec:evaluation_framework}, we propose a methodology for evaluating agents on SWE-Bench-CL. This framework centers on an interactive coding agent, built with LangGraph \cite{langgraph2024} and inspired by the SWE-agent project \cite{yang2024sweagentagentcomputerinterfacesenable}, augmented with a semantic memory module. It was developed to overcome challenges with existing harnesses (Section~\ref{sec:harness_challenges}) for greater transparency in assessing continual learning.
    \item \textbf{Specialized Continual Learning Metrics:} In Section~\ref{sec:metrics}, we define a suite of evaluation metrics tailored for assessing continual learning in software engineering, addressing success rate, tool use efficiency, knowledge transfer, and forgetting.
    \item \textbf{A Rigorous Experimental Protocol:} In Section~\ref{sec:proposed_experiments}, we outline experiments designed to validate the continual learning metrics, comparing memory-enabled/disabled agents and measuring stability–plasticity trade‑offs.
\end{enumerate}

Our goal is to provide the research community with a robust benchmark and a principled evaluation approach to catalyze the development of more adaptive AI agents for software engineering. The codebase is available at \url{https://github.com/thomasjoshi/agents-never-forget/}.

\section{Related Work}
\label{sec:related_work}

The evaluation of LLMs on code has rapidly advanced. Initial benchmarks like HumanEval \citep{chen2021evaluatinglargelanguagemodels} and MBPP \citep{austin2021programsynthesislargelanguage} focused on functional correctness for small problems. CodeXGLUE \citep{lu2021codexgluemachinelearningbenchmark} offered broader tasks. SWE-Bench \citep{jimenez2024swebenchlanguagemodelsresolve} and its verified version \cite{SWEBenchVerified} advanced with realistic GitHub issues but assess isolated tasks. Their harnesses, while powerful for static evaluation, can be misaligned with derived datasets like SWE-Bench-CL, which have different base versions and sequential structures, as discussed in Section~\ref{sec:harness_challenges}.

Continual Learning (CL) aims to enable systems to learn sequentially without catastrophic forgetting \citep{PARISI201954,Delange_2021}. While explored in computer vision, CL in complex generative tasks like software engineering is emerging.

LLM-based agents, such as SWE-agent \cite{yang2024sweagentagentcomputerinterfacesenable}, which use tools and interactive reasoning (e.g., ReAct \citep{yao2023reactsynergizingreasoningacting}), represent a promising direction for tackling complex software tasks. Memory augmentation, especially Retrieval Augmented Generation (RAG) \citep{lewis2021retrievalaugmentedgenerationknowledgeintensivenlp}, is often used to provide external knowledge. Our work uniquely combines these threads by proposing SWE-Bench-CL, a benchmark specifically for evaluating the \emph{continual learning} abilities of such agents in the software engineering domain, particularly through the use of task-history-based semantic memory.

\section{The SWE-Bench-CL Benchmark}
\label{sec:swe_bench_cl_benchmark}

\subsection{Motivation and Design Goals}
Our primary motivation was to create a benchmark that moves beyond static, one-shot evaluation of coding LLMs and instead assesses their ability to learn and adapt over time within the context of evolving software projects. Traditional benchmarks, while valuable, do not capture an agent's capacity to:
\begin{itemize}[nosep]
    \item Accumulate knowledge from previously solved issues.
    \item Transfer learned patterns or solutions to new, related problems (forward transfer).
    \item Retain proficiency on older tasks after learning new ones (resist catastrophic forgetting).
    \item Adapt its problem-solving strategies or tool usage as it gains experience with a codebase.
\end{itemize}
SWE-Bench-CL is designed to directly measure these attributes by structuring tasks sequentially.

\subsection{Dataset Construction}
SWE-Bench-CL is a continual learning adaptation of the SWE-Bench Verified dataset \cite{SWEBenchVerified}. GitHub issues and their corresponding code patches were transformed into a series of learning sequences, each associated with a distinct software repository, designed to simulate a developer's learning trajectory within specific, real-world codebases.

Construction involved selecting repositories with sufficient task instances ($\geq 15$ tasks) from SWE-Bench Verified. For each repository, a learning sequence was created employing several strategies:
\begin{itemize}[nosep,leftmargin=*]
    \item \textbf{Curriculum Learning:} After being primarily ordered by their creation timestamp, tasks are further ordered by difficulty, estimated by human fix time (<15 min, 15 min - 1 hr, 1-4 hr, >4 hr), presenting easier tasks first. This curriculum-based approach is motivated by findings suggesting that training models on tasks of progressively increasing difficulty can lead to better generalization, improved learning efficiency, and enhanced performance \citep{khajehabdollahi2024emergentmechanismslongtimescales, Bengio2009Curriculum}, and facilitate more stable and effective learning when fine-tuning LLMs \citep{long2025generalizationmedicallargelanguage, shi2025efficientreinforcementfinetuningadaptive}.
    \item \textbf{Dependency Awareness:} Modified file paths from ground truth patches (\texttt{patch}) were extracted for each task. This information was used to identify potential dependencies between tasks (i.e., tasks modifying overlapping sets of files) for the future study of knowledge transfer.
\end{itemize}
The resulting dataset comprises 8 sequences from distinct repositories, totaling 273 tasks. Each task includes metadata (repository, base commit, creation date, difficulty), the problem statement, developer hints, evaluation details (ground truth patch, test setup patch, \texttt{FAIL\_TO\_PASS} and \texttt{PASS\_TO\_PASS} test case lists), and continual learning context (sequence position, difficulty score, potential dependencies, modified files). High-level statistics are summarized in Table~\ref{tab:dataset_stats}. The complete dataset is available in JSON format on our GitHub repository.

\begin{table}[ht]
\centering
\caption{\textbf{SWE-Bench-CL Dataset Statistics per Sequence}}
\label{tab:dataset_stats}
\resizebox{\textwidth}{!}{%
\begin{tabular}{@{}lccccccc@{}}
\toprule
Repository & Tasks & Easy (<15m) & Medium (15m-1h) & Hard (1-4h) & Very Hard (>4h) & Tasks w/ Dependencies (\%) \\
\midrule
django/django & 50 & 50 & 0 & 0 & 0 & 25 (50\%) \\
sympy/sympy & 50 & 25 & 25 & 0 & 0 & 12 (24\%) \\
sphinx-doc/sphinx & 44 & 22 & 17 & 4 & 1 & 23 (52\%) \\
matplotlib/matplotlib & 34 & 15 & 19 & 0 & 0 & 13 (38\%) \\
scikit-learn/scikit-learn & 32 & 13 & 18 & 1 & 0 & 4 (13\%) \\
astropy/astropy & 22 & 4 & 15 & 3 & 0 & 3 (14\%) \\
pydata/xarray & 22 & 5 & 15 & 1 & 1 & 13 (59\%) \\
pytest-dev/pytest & 19 & 8 & 8 & 3 & 0 & 7 (37\%) \\
\bottomrule
\end{tabular}%
}
\end{table}

\section{Preliminary Analysis of SWE-Bench-CL: Characteristics and Implications}
\label{sec:preliminary_dataset_analysis}
To better understand the nature of SWE-Bench-CL and the specific challenges it presents for continual learning agents, we conducted preliminary analyses of its structural properties. These analyses inform the design of our proposed evaluation framework and highlight considerations for developing agents capable of continuous learning in software engineering contexts.

\subsection{Low Inter-Task Structural Similarity}
\label{subsec:structural_overlap}
\begin{figure}[htbp]
  \centering
  \includegraphics[width=\linewidth]{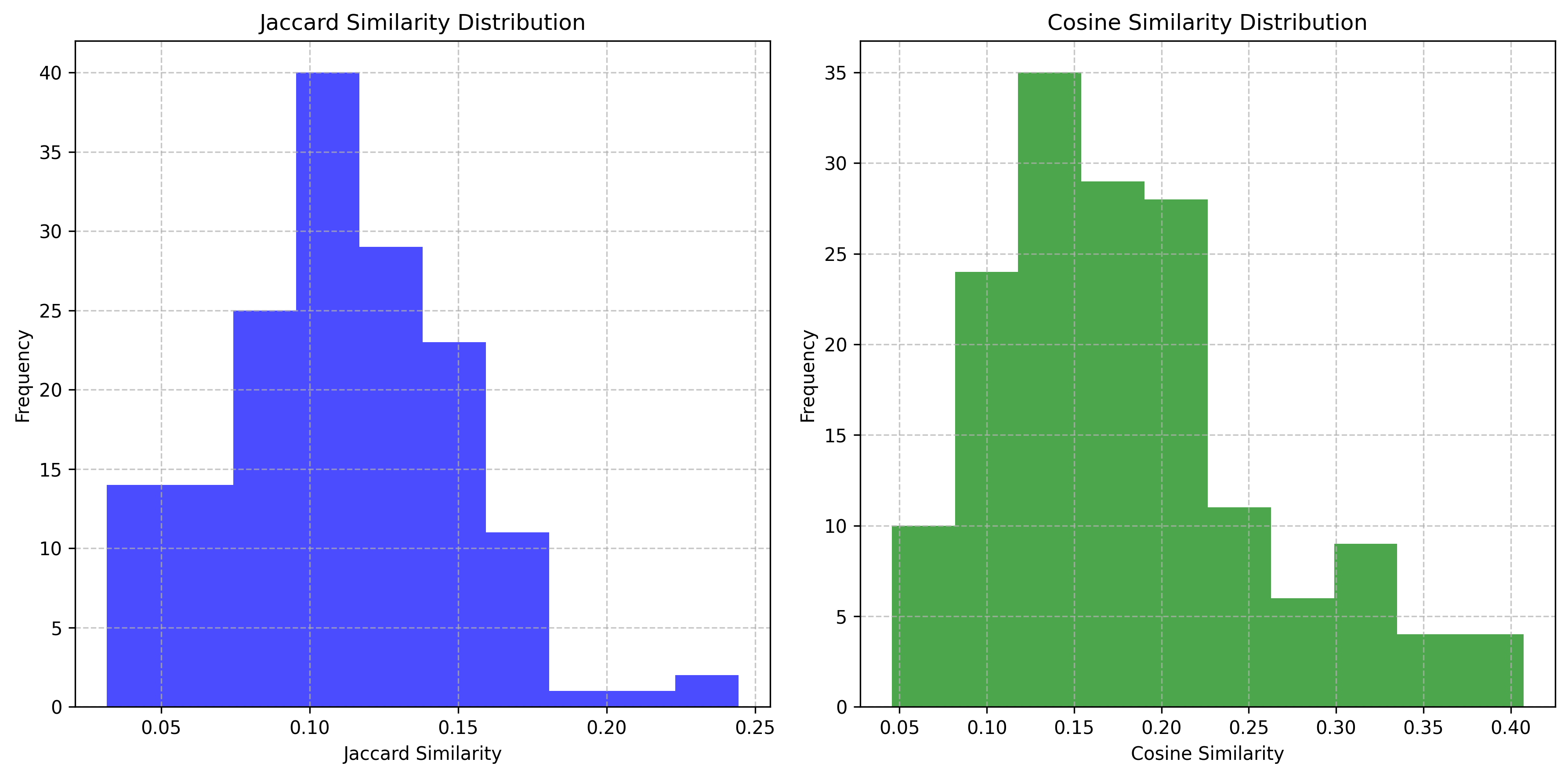}
  \caption{Distribution of Patch-Level Similarity Across Tasks in SWE-Bench-CL. Left: Jaccard similarity. Right: Cosine similarity. Most patch/task pairs exhibit low structural similarity, reinforcing SWE-Bench-CL's role as a high-variance benchmark.}
  \label{fig:patch-similarity-distribution}
\end{figure}

We investigated the structural overlap between the ground truth patches of different tasks within SWE-Bench-CL using Jaccard similarity of token sets and cosine similarity of TF-IDF embeddings (Figure~\ref{fig:patch-similarity-distribution}). Our findings indicate a high degree of independence among most issue-patch pairs. The average Jaccard similarity was $0.1114$ and the average cosine similarity was $0.1792$. Few task pairs exhibited substantial overlap, with only one pair exceeding a cosine similarity of $0.4$. Even when stratified by difficulty, within-group similarities remained modest (e.g., Easy-Easy Jaccard = $0.1225$), and across-group similarities were lower still (Easy-Hard Jaccard = $0.0353$).

\vspace{-4mm}

\textbf{Implications } This low structural redundancy suggests that agents cannot rely on simple surface-level pattern matching between task solutions. It amplifies the challenge of catastrophic forgetting, as knowledge from one task may not be directly reinforced by subsequent, dissimilar tasks. Effective forward transfer will likely depend on learning more abstract problem-solving strategies or leveraging explicit memory of semantically (rather than structurally) similar past experiences. This characteristic validates the need for a benchmark like SWE-Bench-CL that explicitly tests for retention and transfer across distinct problems. It also motivates the inclusion of a semantic memory module in our proposed evaluation agent. While global overlap is low, some repositories (e.g., \texttt{django/django}) showed localized reuse in common modules, suggesting opportunities for intra-repository transfer.

\subsection{Contextual Sensitivity and Prompt Poisoning}
\label{subsec:prompt_poisoning}
To assess how LLM-based agents might be affected by potentially irrelevant contextual information (e.g., from a RAG system retrieving sub-optimal memories), we performed a "prompt poisoning" experiment (Algorithm~\ref{alg:prompt_poisoning}). For a target task $B$, we compared the semantic drift ($1 - \cos(\text{solution}_{\text{clean}}, \text{solution}_{\text{poisoned}})$) in generated solutions when the prompt for $B$ was prepended with an unrelated issue-patch pair from task $A$.

\begin{algorithm}[H]
\caption{Prompt Poisoning Drift Analysis}
\label{alg:prompt_poisoning}
\begin{algorithmic}[1]
\Require SWE-Bench-CL, task difficulty labels, cosine\_similarity(), LLM
\State For $(d_{\text{src}}, d_{\text{tgt}})$ difficulty pairs:
    \State Sample $N$ unrelated task pairs $(A, B)$ with $\text{difficulty}(A) = d_{\text{src}}, \text{difficulty}(B) = d_{\text{tgt}}$.
    \State For each $(A, B)$:
        \State $r_{\text{clean}} \gets \text{LLM}(\text{prompt}(B))$
        \State $r_{\text{poisoned}} \gets \text{LLM}(\text{prompt}(A) + \text{prompt}(B))$
        \State Record $drift \gets 1 - \cos(r_{\text{clean}}, r_{\text{poisoned}})$.
\State Aggregate mean drift.
\end{algorithmic}
\end{algorithm}

\begin{figure}[htbp]
    \centering
    \includegraphics[width=0.8\linewidth]{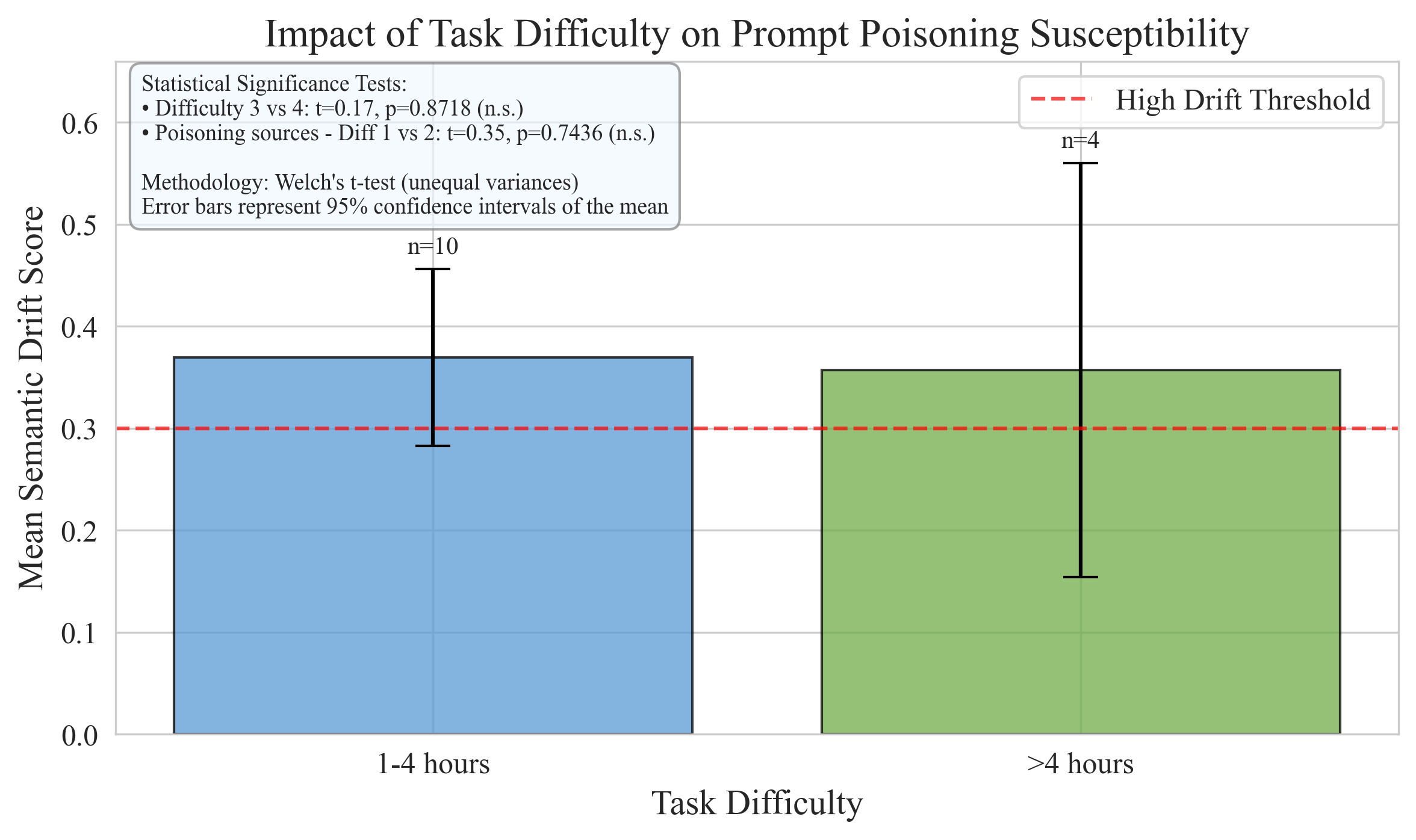}
    \caption{Mean semantic drift induced by prompt poisoning across SWE-Bench-CL tasks, grouped by difficulty of the target task ($1-4$ hours vs. $>4$ hours). Drift is computed as $1 - \cos(\text{clean}, \text{poisoned})$. Red dashed line marks the ``high drift'' threshold ($0.3$). Error bars represent 95\% confidence intervals.}
    \label{fig:drift-by-difficulty}
\end{figure}

As shown in Figure~\ref{fig:drift-by-difficulty}, even structurally dissimilar prompts from easier tasks induced consistently high semantic drift (average $\approx 0.45$) in solutions for more difficult target tasks. While differences between target difficulty groups were not always statistically significant due to sample sizes, the overall high drift indicates that LLM outputs are sensitive to contextual inputs, even if those inputs are not directly relevant to the immediate task.

These findings highlight a vulnerability for memory-augmented agents: naively retrieving and incorporating past experiences can degrade performance if the retrieval mechanism is not highly discerning or if the agent cannot effectively gate irrelevant information. This motivates the need for sophisticated semantic retrieval in our proposed agent's memory system and underscores the importance of evaluating how agents \emph{use} memory, not just whether they have access to it.

\subsection{Informing Agent Design and Fine-Tuning Strategies}
The structural properties of SWE-Bench-CL directly inform design choices for both agentic evaluation frameworks and potential fine-tuning strategies developing continually learning coding LLMs:
\begin{itemize}[nosep,leftmargin=*]
    \item \textbf{Agent Memory Design:} The prevalence of distinct tasks suggests that an effective memory system for agents evaluated on SWE-Bench-CL should prioritize semantic similarity (e.g., issue type, error patterns) over superficial structural similarity for retrieving past experiences.
    \item \textbf{Evaluation of Context Use:} Agentic evaluation frameworks should not only measure task success but also analyze how agents interact with and are influenced by retrieved context, especially given the observed sensitivity to prompt poisoning.
    \item \textbf{Curriculum for Fine-Tuning:} The dataset's inherent curriculum structure (chronological and difficulty-based ordering) provides a natural sequence for fine-tuning LLMs. A fine-tuning protocol could leverage this sequential nature to adapt a base LLM to specific repositories or general software engineering patterns over time, by sequentially fine-tuning on tasks and evaluating for knowledge accumulation and forgetting. This is complementary to the RAG-based memory in our proposed test-time agent.
\end{itemize}
These preliminary analyses confirm that SWE-Bench-CL provides a rich and challenging environment for developing and evaluating agents that aim to continuously learn.

\section{Empirical Evaluation \& Motivation for an Agentic Framework}
\label{sec:harness_challenges}
To empirically ground the need for a specialized evaluation approach, we attempted to evaluate agents on SWE-Bench-CL tasks using the official SWE-Bench evaluation harness \citep{jimenez2024swebenchlanguagemodelsresolve}. This harness provides \textbf{pre-configured Docker "dump containers"} primarily for the original SWE-Bench and SWE-Bench Lite datasets, and does \textit{not} natively support the SWE-Bench Verified subset from which SWE-Bench-CL is derived. We generated patches with a range of LLMs (e.g., \texttt{CodeLlama-13B} \cite{roziere2023code}, \texttt{Gemma-3-12B}, \cite{gemmateam2025gemma3technicalreport}, \texttt{Mistral-7B} \cite{jiang2023mistral7b} via Ollama, and Google's \texttt{Gemini-2.0-Flash}) using a standardized prompt (see Appendix \ref{app:prompts}) under two main conditions: a baseline (independent tasks) and a memory-enabled condition (prompt augmented with semantically retrieved information from previously attempted tasks).

\begin{figure}[htbp]
  \centering
  \begin{minipage}{0.495\linewidth}
    \centering
    \includegraphics[width=\linewidth]{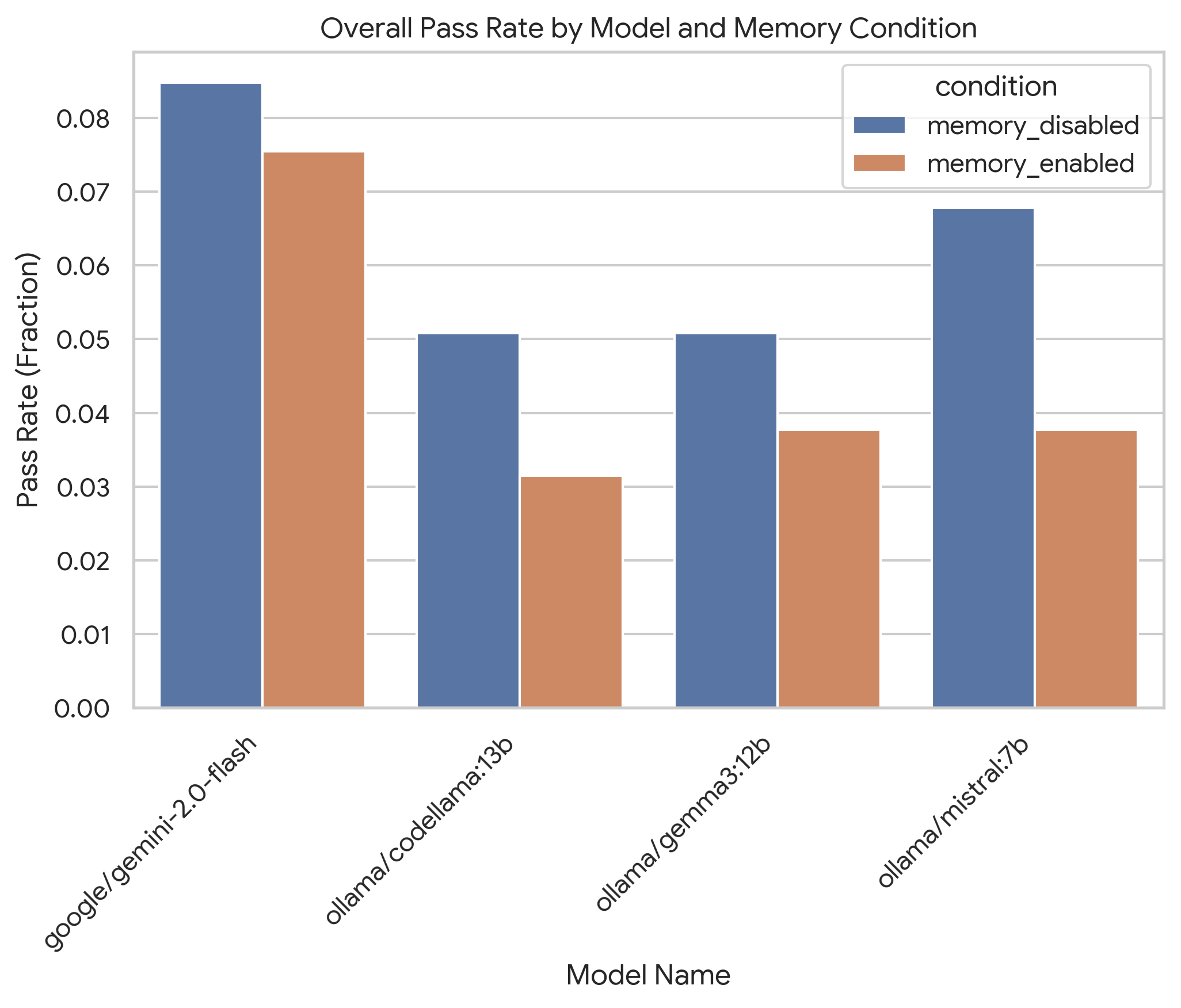}
    \caption{Overall Pass Rate on SWE-Bench-CL tasks using the standard SWE-Bench harness. Low pass rates highlight mismatch with static tooling.}
    \label{fig:harness_pass_rate}
  \end{minipage}\hfill 
  \begin{minipage}{0.495\linewidth}
    \centering
    \includegraphics[width=0.97\linewidth]{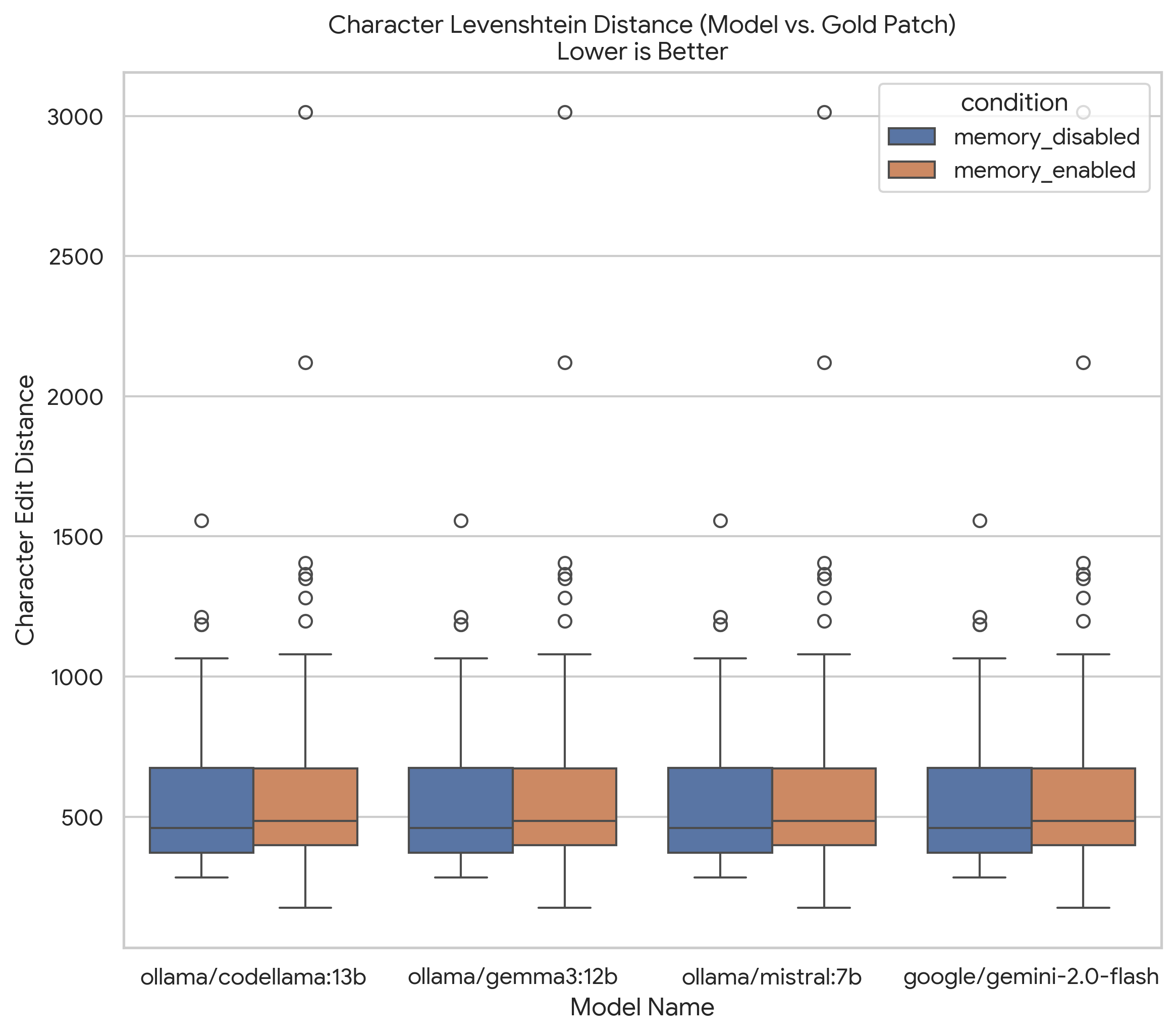}
    \caption{Character Levenshtein Distance (Model vs. Gold Patch) for SWE-Bench-CL tasks. High distances indicate evaluation challenges.}
    \label{fig:harness_levenshtein}
  \end{minipage}
\end{figure}

This endeavor highlighted several critical incompatibilities:
\begin{itemize}[nosep,leftmargin=*]
    \item \textbf{Mismatch of Containers and Task Definitions:} The harness’s Docker containers and associated test execution scripts are tightly coupled to the specific file layouts, patch formats, and codebase states of the original SWE-Bench instances. SWE-Bench-CL, being built upon SWE-Bench Verified and restructured chronologically, often presents tasks where these assumptions no longer hold. Even \textbf{minor differences} can lead to \textbf{runtime errors} or \textbf{silent evaluation failures}.
    \item \textbf{Inconclusive and Unreliable Results:} After adapting the harness for SWE-Bench-CL, performance metrics were extremely low and highly variable, with frequent harness execution issues. As shown in Figure~\ref{fig:harness_pass_rate}, overall pass rates remained consistently low (generally below 8.5\%, often significantly lower), with the memory-enabled condition typically performing on par with or slightly worse. Similarly, Character Levenshtein distances to ground truth patches (Figure~\ref{fig:harness_levenshtein}) were persistently high. This made it impossible to draw reliable conclusions.
\end{itemize}

The generally \textbf{poor performance of the memory-enabled condition} (Figure~\ref{fig:harness_pass_rate}) is plausible: high failure rates due to harness incompatibility likely populated "memory" with information from predominantly failed prior attempts, creating a \textbf{"garbage-in, garbage-out"} scenario.

These findings underscore a fundamental limitation of evaluating evolving, derived benchmarks with static, patch-only frameworks: they cannot flexibly accommodate variations differing from their original target. This critical gap motivated our development of the proposed agentic evaluation framework (Section~\ref{sec:evaluation_framework}), which treats evaluation as an interactive process for a more reliable assessment of problem-solving and learning capabilities on SWE-Bench-CL.

\section{Proposed Agentic Evaluation Framework}
\label{sec:evaluation_framework}
To effectively evaluate performance on the SWE-Bench-CL benchmark, we propose an interactive, agent-based framework integrating semantic memory, inspired by SWE-agent \cite{yang2024sweagentagentcomputerinterfacesenable}.

\subsection{Rationale for a Custom Agent-Based Evaluation Framework}
Evaluating continual learning in software engineering necessitates more than static code generation. Initial attempts to use the standard SWE-bench harness \citep{jimenez2024swebenchlanguagemodelsresolve} with our custom dataset and CL setup faced challenges like opaque error messages, hindering debugging and analysis. To achieve greater transparency, control over agent-environment interaction, and enhanced debuggability for CL experiments—particularly for integrating complex agentic behaviors like semantic memory—we propose a custom agentic framework providing clearer insights into:
\begin{itemize}[nosep,leftmargin=*]
    \item \textbf{Tool Use:} Agent's ability to select and use tools effectively.
    \item \textbf{Multi-Step Reasoning:} Iterative sequences of actions, observations, and plan adjustments.
    \item \textbf{Feedback Incorporation:} Interpreting tool feedback (e.g., linter errors, test failures) and adapting.
    \item \textbf{Simulating Experience:} Agent interacts with the environment and remembers past interactions.
\end{itemize}

\subsection{Agent Architecture and Agent-Computer Interface (ACI)}
Implemented using LangGraph \cite{langgraph2024} for stateful, graph-based execution, the agent is model-agnostic (configurable for OpenAI, Anthropic, Google, and local Ollama models) with standardized generation parameters (e.g., low \texttt{temperature}, \texttt{max\_tokens}). Agent state is tracked in a Pydantic model (\texttt{AgentState}). Drawing inspiration from SWE-agent \cite{yang2024sweagentagentcomputerinterfacesenable}, we developed an Agent-Computer Interface (ACI) and structured interaction patterns, while preserving the LangGraph foundation and integrating our novel semantic memory system.

Based on SWE-agent \cite{yang2024sweagentagentcomputerinterfacesenable}, our ACI provides LLMs with tools to interact with the software environment: \textbf{Navigation/Search} (\texttt{find\_file}, \texttt{search}), \textbf{File Viewing} (\texttt{file\_viewer} with windowed view), \textbf{Editing} (\texttt{edit} tool with \texttt{flake8} linting before applying edits), and \textbf{Execution} (\texttt{run\_tests} for shell commands). These tools accept necessary context from \texttt{AgentState} and return concise, structured, LM-parsable feedback.

\subsection{Semantic Memory System}
\label{subsec:memory}
To explicitly model and evaluate continual learning, we introduce a \texttt{MemorySystem} built using a FAISS \citet{douze2024faiss} vector index. This distinguishes our approach and allows direct comparison of agent performance with/without access to learned experiences.
\begin{itemize}[nosep,leftmargin=*]
    \item \textbf{Storage:} Upon completing a task attempt, key aspects (summaries of problem, solution, rationale; tool usage; success status) are vectorized (e.g., using OpenAI's \texttt{text-embedding-3-small}, \texttt{nomic-embed-text} \cite{nussbaum2025nomicembedtrainingreproducible}) and stored.
    \item \textbf{Retrieval (RAG):} When initiating a new task, the agent queries memory using the current task's problem statement/hints, prioritizing experiences from the same task sequence.
    \item \textbf{Context Integration:} Top-k retrieved memories (including success status, relevance score) are formatted and prepended to the agent's initial prompt for the new task, subject to a configurable token limit (\texttt{max\_context\_tokens}).
\end{itemize}

\subsection{Task Execution Workflow}
For each task in a SWE-Bench-CL sequence, the agent performs:
\begin{enumerate}[label=\arabic*., nosep,leftmargin=*]
    \item \textbf{Repository Setup:} Repository is cloned and/or reset to the task's \texttt{base\_commit}.
    \item \textbf{Agent Interaction:} Using ACI tools and memory, agent iteratively generates \texttt{DISCUSSION}/\texttt{COMMAND} pairs. Commands trigger tools, update \texttt{AgentState}, and return results. Error handling manages failures.
    \item \textbf{Termination:} Loop concludes on \texttt{submit} command, reaching a turn limit, or excessive errors.
\end{enumerate}

\section{Proposed Evaluation Metrics}
\label{sec:metrics}
To assess continual learning capabilities on SWE-Bench-CL, we define metrics for: (i) \emph{solving new issues}, (ii) \emph{retaining prior knowledge}, (iii) \emph{transferring knowledge}, and (iv) \emph{operating efficiently}. We record performance by testing on all previously seen tasks after each new issue, yielding a performance matrix $a_{i,j}$.

\paragraph{Notation} Let \( N \) be total tasks, \( a_{i,j} \) success rate on task \( j \) after training on task \( i \), \( \bar{a}_{0,j} \) zero-shot success on task \( j \).

\begin{tcolorbox}[sharp corners, boxrule=0.5pt, width=\textwidth]
\scriptsize
\renewcommand{\arraystretch}{0.8}
\begin{tabular}{l p{0.38\textwidth} p{0.42\textwidth}}
\toprule
\textbf{Metric} & \textbf{Formula} & \textbf{Definition} \\
\midrule
$\mathrm{SR}_{i,j}$ 
  & $a_{i,j} = \dfrac{p_j^{\mathrm{pass}}}{p_j^{\mathrm{total}}}$ 
  & Per-task success rate: immediate proficiency on task $j$ after training on $i$. \\
$\mathrm{ACC}$ 
  & $\displaystyle \frac{1}{N}\sum_{j=1}^N a_{N,j}$ 
  & Average accuracy after learning all $N$ tasks. \\
$\mathrm{F}$ 
  & $\displaystyle \frac{1}{N-1}\sum_{j=1}^{N-1}\bigl(\!\max_{1\le k\le j}\!a_{k,j}-a_{N,j}\bigr)$ 
  & Average forgetting: performance degradation on earlier tasks. \\
$\mathrm{FT}$ 
  & $\displaystyle \frac{1}{N-1}\sum_{i=1}^{N-1}\bigl(a_{i,i+1}-\bar a_{0,i+1}\bigr)$ 
  & Forward transfer: benefit of prior tasks on new ones (vs.\ zero-shot). \\
$\mathrm{BWT}$ 
  & $\displaystyle \frac{1}{N-1}\sum_{i=1}^{N-1}\bigl(a_{N,i}-a_{i,i}\bigr)$ 
  & Backward transfer: impact of learning new tasks on past tasks. \\
$\mathrm{AULC}$ 
  & $\displaystyle \frac{1}{N}\sum_{i=1}^N\Bigl(\frac{1}{i}\sum_{k=1}^i a_{k,k}\Bigr)$ 
  & Area under learning curve: integrated performance over training steps. \\
$\mathrm{TUE}$ 
  & $\dfrac{\text{median}(\text{time}\mid\text{success})}
         {\text{median}(\text{time}\mid\text{all})}$ 
  & Tool-use efficiency: ratio of median execution times for successful vs.\ all runs. \\
\bottomrule
\end{tabular}
\end{tcolorbox}

\paragraph{Composite CL-Score:}  
Combines metrics into a single score with tunable weights $\lambda$:
\[
  \mathrm{CL\text{-}Score} = \mathrm{ACC}
    - \lambda_{F}\,F
    + \lambda_{FT}\,\mathrm{FT}
    + \lambda_{BWT}\,\mathrm{BWT}
    + \lambda_{AULC}\,\mathrm{AULC}
    + \mathrm{CL\!-\!F}_{\beta},
    \quad
    \lambda \ge 0
\]
Continual‑Learning-F$_\beta$ score (\(\mathrm{CL\!-\!F}_{\beta}\)) is defined in Section \ref{sec:cl-f1}.

\paragraph{Role of the \(\lambda\) Weights:}
All $\lambda$ coefficients may be adjusted to reflect relative priorities:
\begin{tcolorbox}[sharp corners, boxrule=0.5pt, width=\textwidth]
\scriptsize
\renewcommand{\arraystretch}{1.3}
\begin{tabular}{l p{0.20\textwidth} p{0.65\textwidth}}
\toprule
\textbf{Factor} & \textbf{Name} & \textbf{Description and Use Case} \\
\midrule
$\lambda_{\mathrm{F}}$ 
  & \textbf{Forgetting Penalty}
  & Increase for high memory retention (e.g., safety-critical code). \\
$\lambda_{\mathrm{FT}}$ 
  & \textbf{Forward Transfer Reward}
  & Increase for rapid adaptation to new functions (e.g., rapid-prototyping). \\
$\lambda_{\mathrm{BWT}}$ 
  & \textbf{Backward Transfer Reward}
  & Increase if later tasks should improve earlier ones (e.g., modular libraries). \\
$\lambda_{\mathrm{AULC}}$ 
  & \textbf{Learning Speed Factor}
  & Modulates initial learning speed vs. eventual proficiency. Increase for fast competence. \\
$\lambda_{\mathrm{TUE}}$ 
  & \textbf{Tool-Use Efficiency}
  & Increase to prioritize resource efficiency (fewer tool calls, less time) for practical deployment. \\
\bottomrule
\end{tabular}
\end{tcolorbox}
The specific values for $\lambda$ should be determined based on research questions or deployment goals.

\subsection{Continual‑Learning F1 (CL‑F1)}
\label{sec:cl-f1}
The core challenge in CL is the \textit{stability-plasticity dilemma} \citep{grossberg1982does, mermillod2013stability}: an agent must be plastic enough to learn new information effectively, yet stable enough to prevent new learning from catastrophically disrupting previously acquired knowledge. We propose the Continual Learning F1-Score (CL-F1) to explicitly quantify this trade-off. 

We define two components:
\begin{itemize}[nosep,leftmargin=*]
  \item \textbf{CL-Plasticity (CL-P): Immediate Proficiency.} Measures the agent's ability to learn and correctly solve new tasks. Defined as the average success rate on each task $i$ immediately after it is processed: $$
          \mathrm{CL\text{-}P} = \frac{1}{N}\sum_{i=1}^{N} a_{i,i}
          $$
  \item \textbf{CL-Stability (CL-S): Knowledge Retention.} Measures the agent's ability to retain performance on previously learned tasks. Defined as one minus Average Forgetting (F):
        \[
          \mathrm{CL\text{-}S} = 1 - \mathrm{F} = 1 - \frac{1}{N-1}\sum_{j=1}^{N-1} \Bigl(\max_{1\le k \le j}a_{k,j} - a_{N,j}\Bigr)
        \]
\end{itemize}
The \textbf{CL-F1 score} is the harmonic mean of CL-Plasticity and CL-Stability:
\[
  \boxed{\displaystyle
    \mathrm{CL\text{-}F1}
    =
    2 \cdot \frac{\mathrm{CL\text{-}P} \times \mathrm{CL\text{-}S}}
         {\mathrm{CL\text{-}P} + \mathrm{CL\text{-}S}}
  }
  \qquad
  (0 \le \mathrm{CL\text{-}F1} \le 1)
\]

\paragraph{Interpretation and Rationale.} A high CL-F1 score is achieved when the agent demonstrates both strong immediate learning (high CL-P) and robust retention (high CL-S). The harmonic mean penalizes imbalance. This metric addresses the stability-plasticity dilemma by rewarding a balance, adapting F-scores to sequential learning \citep[cf.][]{Chaudhry_2018,lopezpaz2022gradientepisodicmemorycontinual}.

\paragraph{Generalized CL-F$_\beta$ Score.} We introduce the CL-F$_\beta$ score:
\[
  \boxed{\displaystyle
    \mathrm{CL\text{-}F}_{\beta}
    =
    (1+\beta^{2})\,
    \frac{\mathrm{CL\text{-}P} \times \mathrm{CL\text{-}S}}
         {\beta^{2}\,\mathrm{CL\text{-}P} + \mathrm{CL\text{-}S}}
  }
  \qquad
  (\beta > 0)
\]
The parameter \(\beta\) controls weighting: \(\beta = 1\) (balanced CL-F1), \(0 < \beta < 1\) (emphasizes plasticity), \(\beta > 1\) (emphasizes stability). The choice of \(\beta\) depends on task requirements; \(\beta=1\) is standard for general benchmarking.

\section{Proposed Experiments and Hypotheses}
\label{sec:proposed_experiments}
The CL metrics (Section~\ref{sec:metrics}) provide the quantitative foundation for experiments crucial for validating SWE-Bench-CL and our proposed metrics. While full empirical evaluation using our agentic framework is ongoing, key experiments will:

Compare memory-enabled versus memory-disabled agents using various LLMs (e.g., GPT-4o, Claude 3.7 Sonnet, open models like DeepSeek-V3 \cite{deepseekai2025deepseekv3technicalreport} or Gemma3-27B \cite{gemmateam2025gemma3technicalreport}) across multiple SWE-Bench-CL sequences. Key analyses include:
\begin{itemize}[nosep,leftmargin=*]
    \item \textbf{Impact of Semantic Memory:} Quantify performance gains (ACC, SR$_{i,j}$), forward transfer (FT), and tool-use efficiency (TUE) due to the semantic memory system. It is hypothesized that memory will improve overall accuracy, yield positive forward transfer, and enhance tool-use efficiency.
    \item \textbf{Stability-Plasticity Trade-off:} Assess average forgetting (F), CL-Plasticity (CL-P), CL-Stability (CL-S), and the composite CL-F$_\beta$ score. It is hypothesized that memory-augmented agents will exhibit higher CL-Stability (lower forgetting) and achieve a better CL-F$_\beta$ score, demonstrating an improved balance between learning new tasks and retaining old knowledge.
\end{itemize}

\section{Limitations}
\label{sec:limitations}
While SWE‑Bench‑CL advances continual‑learning benchmarks for coding agents, it has limitations:
\begin{itemize}[nosep,leftmargin=*]
  \item \textbf{Repository and Language Scope.} Evaluates only 8 open‑source Python repositories, each with at most 50 issues per sequence. This scope may not generalize to other languages, larger projects, or development workflows with branching, parallel issue resolution, and pull‑request reviews.
  \item \textbf{Coarse Difficulty and Dependency Signals.} Task difficulty is estimated by human fix‑time categories, which are subjective. Dependency awareness relies on overlapping file paths, potentially missing deeper semantic or API‑level dependencies.
  \item \textbf{Static Sequence Assumptions.} Issues are presented strictly in chronological order with a built‑in curriculum. Real codebases often interleave unrelated issues, hotfixes, and refactorings; our simplified sequence may not reflect true software evolution dynamics.
\end{itemize}

\section{Conclusion}
\label{sec:conclusion}
The long-term vision is to cultivate coding agents that not only address immediate software engineering challenges but also continuously augment their expertise and reliability through sustained interaction with evolving codebases and task streams, truly embodying the principle of "agents that continuously learn."

\newpage
\appendix
\section{Appendix: Prompt Template for Patch Generation (SWE-Bench Harness Experiments)}
\label{app:prompts}

Adapted from \cite{jimenez2024swebenchlanguagemodelsresolve}, the following prompt template was used as the basis for generating patches in the experiments described in Section~\ref{sec:harness_challenges}. Placeholders like \texttt{problem\_statement}, \texttt{retrieved\_context}, etc., were filled dynamically based on the task and experimental condition (e.g., \texttt{retrieved\_context} would contain information from past tasks for the memory-enabled condition). The \texttt{patch\_example\_content} was a static example of a diff.

\begin{verbatim}
You will be provided with a partial code base and an issue statement explaining 
a problem to resolve.
The goal is to generate a code patch in the **unified diff format** that 
resolves the issue.

<issue>
{problem_statement}
</issue>

Relevant context from the repository ({repo} at commit {base_commit}):
<code>
{retrieved_context}

**Hints (if any from the original issue):**
{hints_text}

**Files to consider (based on gold solution, try to identify which files to modify):**
{text_files}
</code>

Here is an example of a patch file. It consists of changes to the codebase. 
It specifies the file names, the line numbers of each change, and the removed 
and added lines. A single patch file can contain changes to multiple files. 

<patch>
--- a/file.py
+++ b/file.py
@@ -1,27 +1,35 @@
 def euclidean(a, b):
-    while b:
-        a, b = b, a % b
-    return a
+    if b == 0:
+        return a
+    return euclidean(b, a % b)
 
 
 def bresenham(x0, y0, x1, y1):
     points = []
     dx = abs(x1 - x0)
     dy = abs(y1 - y0)
-    sx = 1 if x0 < x1 else -1
-    sy = 1 if y0 < y1 else -1
-    err = dx - dy
+    x, y = x0, y0
+    sx = -1 if x0 > x1 else 1
+    sy = -1 if y0 > y1 else 1
 
-    while True:
-        points.append((x0, y0))
-        if x0 == x1 and y0 == y1:
-            break
-        e2 = 2 * err
-        if e2 > -dy:
+    if dx > dy:
+        err = dx / 2.0
+        while x != x1:
+            points.append((x, y))
             err -= dy
-            x0 += sx
-        if e2 < dx:
-            err += dx
-            y0 += sy
+            if err < 0:
+                y += sy
+                err += dx
+            x += sx
+    else:
+        err = dy / 2.0
+        while y != y1:
+            points.append((x, y))
+            err -= dx
+            if err < 0:
+                x += sx
+                err += dy
+            y += sy
 
+    points.append((x, y))
     return points
</patch>

I need you to solve the provded issue by generating a single patch file that I 
can apply directly to this repository using git apply. Please respond with a 
single patch file in the format shown above.

Respond below:
\end{verbatim}

\end{document}